\documentclass[10pt,twocolumn,letterpaper]{article}
\usepackage[pagenumbers]{cvpr} 

\usepackage{multirow}

\usepackage{amsmath}
\usepackage{amssymb}
\usepackage{mathtools}
\usepackage{amsthm}

\theoremstyle{plain}
\newtheorem{theorem}{Theorem}[section]
\newtheorem{proposition}[theorem]{Proposition}
\newtheorem{lemma}[theorem]{Lemma}

\theoremstyle{definition}

\newtheorem{assumption}[theorem]{Assumption}
\theoremstyle{remark}

\definecolor{aquatic}{HTML}{DA5C54}
\definecolor{birds}{HTML}{56D3DB}

\definecolor{cvprblue}{rgb}{0.21,0.49,0.74}
\usepackage[pagebackref,breaklinks,colorlinks,allcolors=cvprblue]{hyperref}

\usepackage[capitalize,noabbrev]{cleveref}
\crefname{section}{Sec.}{Secs.}
\Crefname{section}{Section}{Sections}
\Crefname{table}{Table}{Tables}
\crefname{table}{Tab.}{Tabs.}
\crefname{equation}{Eq.}{Eqs.}
\Crefname{equation}{Equation}{Equations}
\crefname{theorem}{Prop.}{Props.}
\Crefname{theorem}{Proposition}{Propositions}

\def\paperID{*****} 
\def\confName{CVPR}
\def\confYear{2025}

\title{Neutralizing Token Aggregation via Information Augmentation \\ for Efficient Test-Time Adaptation}

\author{Yizhe Xiong$^{1,2\ *}$\quad Zihan Zhou$^{1,2\ *}$\quad Yiwen Liang$^{1,2\ *}$\quad Hui Chen$^{2}$\quad Zijia Lin$^{1}$\quad Tianxiang Hao$^{1}$\\ Fan Zhang$^{1}$\quad Jungong Han$^{3}$\quad Guiguang Ding$^{1,2}$\\
$^1$School of Software, Tsinghua University\  $^2$BNRist, Tsinghua University\\  $^3$Department of Automation, Tsinghua University\\
{\tt\small xiongyizhe2001@gmail.org}
}

\begin{document}
\maketitle
\def\thefootnote{*}\footnotetext{These authors contributed equally to this work}\def\thefootnote{\arabic{footnote}}
\begin{abstract}
Test-Time Adaptation (TTA) has emerged as an effective solution for adapting Vision Transformers (ViT) to distribution shifts without additional training data. 
However, existing TTA methods often incur substantial computational overhead, limiting their applicability in resource-constrained real-world scenarios. 
To reduce inference cost, plug-and-play token aggregation methods merge redundant tokens in ViTs to reduce total processed tokens.
Albeit efficient, it suffers from significant performance degradation when directly integrated with existing TTA methods.
We formalize this problem as Efficient Test-Time Adaptation (ETTA), seeking to preserve the adaptation capability of TTA while reducing inference latency.
In this paper, we first provide a theoretical analysis from a novel mutual information perspective, showing that token aggregation inherently leads to information loss, which cannot be fully mitigated by conventional norm-tuning-based TTA methods.
Guided by this insight, we propose to \textbf{N}eutralize Token \textbf{A}ggregation \textbf{v}ia \textbf{I}nformation \textbf{A}ugmentation (\textbf{NAVIA}).
Specifically, we directly augment the [CLS] token embedding and incorporate adaptive biases into the [CLS] token in shallow layers of ViTs.
We theoretically demonstrate that these augmentations, when optimized via entropy minimization, 
recover the information lost due to token aggregation. 
Extensive experiments across various out-of-distribution benchmarks demonstrate that NAVIA significantly outperforms state-of-the-art methods by over 2.5\%, while achieving an inference latency reduction of more than 20\%, effectively addressing the ETTA challenge.
\end{abstract}    
\section{Introduction}
\label{sec:intro}

Vision transformers (ViT) \cite{vit,DeiT} have significantly advanced the field of computer vision, bringing substantial advancements across various tasks \cite{DETR,SAM}.
Typically, ViTs are evaluated on data drawn from the same distribution as the training dataset. However, real-world applications \cite{yang2024geometry,TaD,SPT} often present data exhibiting a \textit{distribution shift} \cite{torralba2011unbiased}, which can differ markedly from the original training distribution. 
To address those practical challenges, Unsupervised Domain Adaptation (UDA) \cite{DAN,sun2019unsupervised,wilson2020survey,cvisdit,lyu2024learn} has been proposed to learn domain-invariant image representations from unlabeled data drawn from the target distribution.
Nonetheless, UDA methods typically require extra training, leading to inefficiency and limited deployment flexibility.
Recently, Test-Time Adaptation (TTA) \cite{Tent,EATA,SAR,FOA,TPT} has emerged as an alternative, 
adapting models on each test data group immediately after inference via parameter-efficient fine-tuning \cite{Tent,SAR,TPT} or gradient-free update strategies \cite{LAME,FOA,TDA,ZERO,zhou2025bayesian,iTaD,ReTA}.
Compared to traditional UDA, TTA eliminates the entire training stage, rendering it particularly suitable for real-world scenarios.

\begin{figure}
    \centering
    \includegraphics[width=0.92\linewidth]{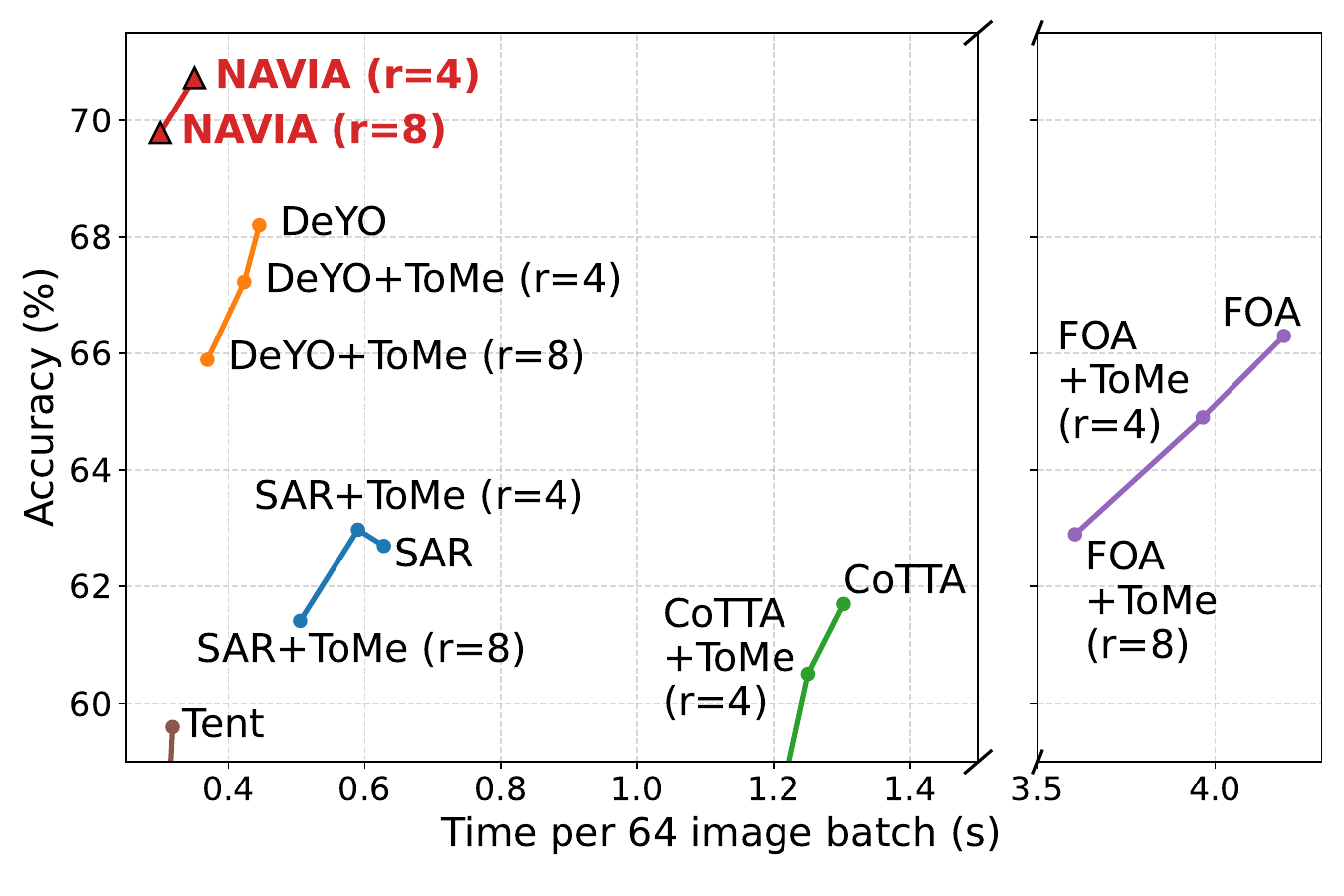}
    \caption{Comparisons between our proposed NAVIA and other ETTA methods in terms of latency-accuracy trade-offs. We apply ToMe with varying token reduction counts ($r$) to existing TTA methods to obtain their ETTA variants.} 
    \label{fig:teaser}
\end{figure}

Although achieving promising out-of-distribution (OOD) performance, current TTA methods generally incur substantial computational overhead by either repeatedly forwarding the same input multiple times \cite{SAR,FOA} or applying different augmentations to the same test batch in successive forward passes \cite{TPT,DeYO,ZERO}. 
For instance, SAR \cite{SAR} computes second-order gradients across multiple inference cycles, and DeYO \cite{DeYO} forwards the same test sample using varying patch sequences. 
Such redundancy significantly exacerbates their computational burden during inference, resulting in a computational complexity that is several times higher than that of a standard forward pass. 
As emphasized by \cite{LAME}, the \textit{forward pass in inference} represents the primary efficiency bottleneck in TTA. 
Consequently, prior TTA methods typically overlook efficiency constraints critical for practical deployment, severely restricting their applicability on edge devices with constrained computational resources.

To address those limitations, we focus on the challenge of \textbf{Efficient Test-Time Adaptation} (\textbf{ETTA}), aiming to improve the \textit{inference efficiency} of TTA methods during inference.
Prior studies have extensively explored model compression methods, including structural compression \cite{llmpruner,yu2022unified,smoothquant} and token aggregation \cite{EViT,ToMe,Tofu}, to enhance inference efficiency in ViTs. 
While model compression methods typically require extensive retraining with substantial data and computational resources to recover the performance lost during compression, token aggregation methods can be adopted for TTA in a plug-and-play manner as they do not alter ViT parameters and do not interfere with the optimization process within TTA.
Naturally, a straightforward solution for ETTA would be a combination of representative TTA approaches, namely Tent \cite{Tent}, FOA \cite{FOA}, SAR \cite{SAR}, DeYO \cite{DeYO}, and CoTTA \cite{cotta} with a parameter-free token aggregation method, ToMe \cite{ToMe}\footnote{In Section 4, we demonstrate that ToMe serves as a neat and strong choice for token aggregation methods.}. 
As shown in \cref{fig:teaser}, even at a modest compression rate (reducing $r=4$ tokens in each layer, $\approx$10\% FLOPs reduction), existing TTA methods suffer significant performance loss. 
Further increasing the compression rate exacerbates this issue, highlighting that a direct combination of existing methods cannot effectively address the efficiency-performance trade-offs central to ETTA.

We further investigate the underlying issues for the failure of existing approaches. 
Through theoretical analysis, 
we demonstrate that existing token aggregation methods inherently result in mutual information loss. 
Additionally, 
standard norm-tuning strategies employed in TTA methods cannot adequately compensate for such loss. 
To effectively mitigate these issues, we propose to \textbf{N}eutralize Token \textbf{A}ggregation \textbf{V}ia \textbf{I}nformation \textbf{A}ugmentation (\textbf{NAVIA}). 
Specifically, we fine-tune the [CLS] embedding alongside norm-tuning during TTA, and 
introduce fine-tunable bias vectors to [CLS] tokens in the shallow layers of the ViT.
We theoretically demonstrate that under the entropy optimization objective, these measures could augment the information capacity reduced by token aggregation, resulting in higher input information during TTA.
Extensive experiments (see \cref{sec:experiments} and \cref{fig:teaser}) across 4 OOD benchmarks demonstrate that NAVIA consistently outperforms existing methods by over 2.5\%, while achieving an inference latency reduction of more than 20\%, effectively addressing the ETTA challenge.

Overall, our contributions are summarized as follows:
\begin{itemize}
    \item We introduce a novel challenge termed Efficient Test-Time Adaptation (ETTA), emphasizing the simultaneous optimization of inference efficiency and prediction performance for test-time adaptation.
    \item Through theoretical analysis from a novel mutual information perspective, we show that simply applying token aggregation methods to TTA for efficiency introduces information loss which cannot be compensated by norm-tuning adopted in existing TTA methods.
    \item 
    Guided by theoretical analysis, we propose NAVIA. NAVIA optimizes the [CLS] token embedding and introduces learnable biases to the [CLS] token in shallow network layers to compensate information loss.
    \item Extensive experiments show that NAVIA significantly outperforms existing methods, improving both inference efficiency and prediction performance. 
    Notably, our method effectively addresses the ETTA challenge for the first time, yielding a better accuracy-efficiency balance.
\end{itemize}

\section{Related Works}
\label{sec:related}

\subsection{Test-Time Adaptation}

Test-Time Adaptation (TTA) enables models to adapt to evolving environments exhibiting distribution shifts during deployment~\cite{insearchtta}. TTA methods generally fall into two categories~\cite{comprehensivesv}: instance-level and batch-level. Instance-level adaptation leverages zero-shot priors from vision-language models via prompt tuning or memory retrieval~\cite{TPT,TDA}, but requires repeated updates or substantial storage, limiting their practicality in resource-constrained environments.
Batch-level methods primarily rely on normalization-based approaches, updating normalization statistics with the current test data~\cite{Tent,DeYO,SAR,FOA}. For instance, Tent~\cite{Tent} minimizes entropy over batches, EATA~\cite{EATA} selects diverse and reliable samples, and SAR~\cite{SAR} employs sharpness-aware minimization to reduce noisy gradients. However, batch-level approaches often demand significant computational resources, restricting their use in real-time or edge-device scenarios. To address these limitations, we pioneer the application of token aggregation in batch-level TTA, significantly enhancing computational efficiency while preserving model adaptability.

\subsection{Model Compression}

Model compression has been widely explored to enhance inference efficiency, particularly for deployment in resource-limited environments. Two prominent approaches are structural compression and token aggregation. 
Structural compression reduces model complexity by directly compressing redundant structures of the model backbone, including model compression \cite{chen2021chasing,llmpruner,yu2022unified,DSMoE,vtclfc}, knowledge distillation \cite{hinton2015distilling,ahn2019variational,tung2019similarity,habib2024knowledgedistillationvisiontransformers}, and model quantization \cite{courbariaux2015binaryconnect,lin2015neural,lin2017towards,cai2020zeroq,vtptq,smoothquant}.
However, these methods typically require extensive retraining, which is infeasible in TTA due to the limited availability of test-time data. 
Token aggregation \cite{diffrate,ToMe}, in contrast, enhances inference efficiency by merging redundant tokens during inference without retraining. Notable methods include ToMe~\cite{ToMe}, which progressively merges similar tokens using lightweight bipartite matching, EViT~\cite{EViT}, which utilizes attention scores to select critical tokens, and ToFu~\cite{Tofu}, which combines pruning and aggregation across layers. 
Despite their effectiveness, token aggregation methods still suffer from significant information loss when directly integrated into TTA.
We conduct a theoretical analysis to address this issue and propose augmenting mutual information, leading to a new state-of-the-art ETTA approach.
\section{Methodology}
\label{sec:methodology}

\subsection{Preliminaries}
\label{sec:prelim}

\noindent\textbf{ViT Model.}
In this paper, we mainly focus on conducting ETTA with \textit{Pre-Norm} Vision Transformer (ViT) models. ViT architectures are transformer-based encoder models, where each encoder consists of a Multi-Head Self-Attention (MHSA) module and a Feed-Forward Network (FFN). In the forward pass, an input image $\mathbf{x}$ is reshaped and projected to $N$ tokens $\mathbf{T}^{\mathrm{Em}}_\mathrm{img}=[t^{\mathrm{Em}}_1,t^{\mathrm{Em}}_2,\cdots,t^{\mathrm{Em}}_N]$\footnote{``Em'' denotes ``Embedding''.}, each with a hidden dimension $d$. A classification ([CLS]) token $t^{\mathrm{Em}}_\mathrm{CLS}$ is prepended to the token sequence, resulting in the ViT model input: $\mathbf{T}^\mathrm{Em}= [t^{\mathrm{Em}}_{\mathrm{CLS}}, t^{\mathrm{Em}}_1, t^{\mathrm{Em}}_2, \cdots, t^{\mathrm{Em}}_N]$.
Inside each encoder block, this token sequence is sequentially processed by the MHSA and the FFN.
Specifically, the MHSA module employs fully-connected (FC) layers to transform input tokens into $Q$, $K$, and $V$ matrices. The self-attention is then carried out via $\mathrm{Softmax}(\frac{QK^T}{\sqrt{D}})V$, and the resulting output is further projected by another FC layer.
Subsequently, the FFN processes the sequence through two FC layers.

\noindent\textbf{Token Aggregation.} 
Token aggregation methods \cite{ToMe,EViT,Tofu,PYRA} are \textit{training-free}, \textit{plug-and-play} approaches that improve the inference efficiency of ViTs by reducing the number of image tokens. 
Those methods are orthogonal to the ViT structure and are capable of flexibly changing the compression rate.
Let $N_l$ and $N_{l+1}$ denote the input and output sequence lengths of image tokens for layer $l$ and $l+1$, respectively, and let $\mathbf{x}_l^{\mathrm{In}}$ and $\mathbf{x}_l^{\mathrm{Out}}$ represent the corresponding input and output token sequences for the token aggregation module at layer $l$.
Specifically, token aggregation methods separate the token sequence into the [CLS] token $t_\mathrm{CLS}$ and the image token sequence $\mathbf{T}^{\mathrm{In}}_{l}=[t_1^l,t_2^l,\cdots,t_N^l]$.
During the aggregation step, existing methods utilize different metrics $g(\cdot)$ to evaluate tokens within $\mathbf{T}^{\mathrm{In}}_{l}$. For instance, ToMe \cite{ToMe} applies similarity-based merging, while EViT
\cite{EViT} applies attentive-token preservation. Based on those evaluation metrics, an aggregation matrix $P_l=g(\mathbf{T}^{\mathrm{In}}_l,t_\mathrm{CLS})\in\mathbb{R}^{N_{l+1}\times N_l}$ is computed and applied to obtain the aggregated output token sequence $\mathbf{T}^{\mathrm{Out}}_{l}=P_l\mathbf{T}_l^{\mathrm{In}}$. Finally, the [CLS] token $t^l_{\mathrm{CLS}}$ is prepended to $\mathbf{T}^{\mathrm{Out}}_{l}$, forming the input for subsequent modules in ViT.

\begin{figure*}
    \centering
    \includegraphics[width=0.82\linewidth]{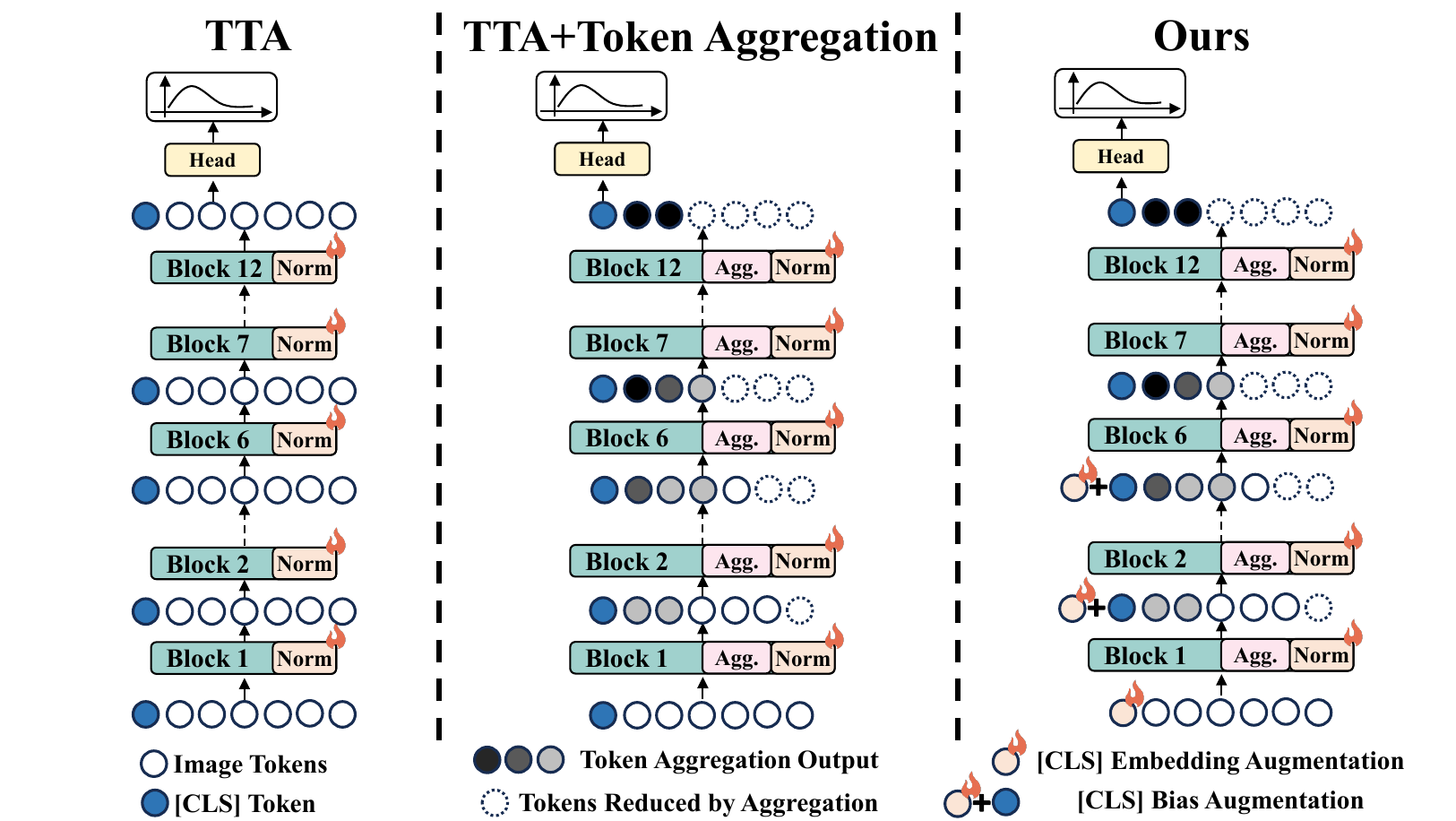}
    \caption{Pipeline comparison between standard TTA methods, ETTA via token aggregation, and our proposed method. ``Agg.'' refers to token aggregation, and ``Norm'' refers to LayerNorms in ViTs that are tunable during TTA. We employ [CLS] embedding augmentation and shallow layer [CLS] bias augmentation to neutralize negative effects caused by token aggregation.}
    \label{fig:pipeline}
\end{figure*}

\subsection{Theoretical Analysis}

Experimental evidence in \cref{fig:teaser} indicate that straightforwardly combining token aggregation with existing TTA approaches leads to significant performance degradation.
Although some TTA methods use norm-tuning to facilitate adaptation, they still fail to compensate for the performance loss caused by token aggregation despite its efficiency.
To delve deeper into this issue, we conduct a theoretical analysis focusing on information loss, revealing that \textit{standard LayerNorm tuning in ViTs used in these methods is insufficient to compensate for the information loss caused by token aggregation}.
In the field of domain adaptation, such information loss can be quantified by the mutual information between image input and the ground-truth label \cite{PCS,cvisdit}.
Prior studies \cite{PCS,cvisdit} have demonstrated that maximizing mutual information \cite{shannon1948mathematical} significantly improves prediction performance under unsupervised conditions. Motivated by these findings, we adopt mutual information ($I(\cdot;\cdot)$) between the ground-truth label $Y$ and input image embeddings as the primary information measure in our analysis.
From the definition of mutual information, we make the following reasonable assumption:

\begin{assumption}
For an image embedding $\mathbf{x}$ and its corresponding ground-truth $Y$, a higher value of $I(\mathbf{x};Y)$ indicates better encoding quality of $\mathbf{x}$, which is more desirable.
\end{assumption}

First, we demonstrate that token aggregation leads to a strict loss of mutual information. Let $f_{\theta;g;\mathbf{x}}(\cdot)$ represent the forwarding process with the pre-trained ViT model $\theta$, token aggregation with metric $g(\cdot)$ and input image $\mathbf{x}$. We use $f_{\theta;I;\mathbf{x}}(\cdot)$ as the forwarding process without token aggregation, as the metric $g(\cdot)$ always returns the identity matrix.

\begin{proposition}
\label{prop:1}
After applying token aggregation, the output [CLS] token contains strictly less information about the label $Y$: 
$I(f_{\theta;g;\mathbf{x}}(t^{\mathrm{Em}}_{\mathrm{CLS}});Y)<I(f_{\theta;I;\mathbf{x}}(t^{\mathrm{Em}}_{\mathrm{CLS}});Y).$
\end{proposition}

\cref{prop:1} indicates that simply applying training-free token aggregation leads to a strict degradation in model predictions. To address this degradation, it is essential to compensate for the resulting information loss. While prominent TTA methods \cite{Tent, SAR, DeYO} have successfully fine-tuned the normalization layers for effective TTA, we show that \textit{norm-tuning alone does not compensate for the information loss} in ViT models\footnote{Current ViTs employ the Pre-Norm structure \cite{vit, UniAttn}, which is adopted in our analysis.}.

\begin{proposition}
\label{prop:2}
    In Pre-Norm ViT models, LayerNorm $(\mathrm{LN}(\cdot))$ tuning cannot compensate for the mutual information loss caused by token aggregation.\footnote{Please refer to the supplementary material for detailed proofs.}
\end{proposition}

\subsection{NAVIA}
\label{sec:our_method}

\cref{prop:2} has shown that simply applying token aggregation to existing TTA methods introduces an information loss that cannot be compensated by subsequent operations.
Accordingly, the optimization should occur \textit{prior to applying the token aggregation matrix}.
This implies that tuning should be performed \textit{directly within the tokens}, a scenario naturally suited to prompt tuning.
Motivated by recent advances in prompt tuning \cite{VPT,VQT,TPT}, we address this issue at the input stage, especially during the construction of input token embeddings within each encoder block. 
However, given that only light optimizations are feasible in TTA, prepending randomly initialized learnable prompts requires intensive tuning to locate optimal embeddings \cite{FOA}, or even requires training an additional processing module to generate effective prompts \cite{liu2024gpt}.
Such an approach introduces significant computational overhead during inference, which is unsuitable for ETTA. 
Consequently, for ETTA, prompt tuning should leverage existing, thoroughly pre-trained prompts, pointing directly to the [CLS] token.

\noindent\textbf{[CLS] Embedding Augmentation.}
Prior studies \cite{tang2024learning,zou2024a} have shown that the [CLS] token can effectively encode domain-specific information beneficial for cross-domain learning. Inspired by those findings, we propose to leverage the [CLS] token to \textit{encode domain-specific compensation information specifically targeted at addressing the information loss induced by token aggregation}. We refer it as \textbf{information augmentation}, serving as a complementary mechanism to norm-tuning. We theoretically demonstrate that optimizing the [CLS] token via entropy minimization compensates for information loss:

\begin{proposition}
\label{prop:3}
Let $\delta$ represent a learned bias vector for the $\mathrm{[CLS]}$ token optimized through entropy minimization. The introduction of $\delta$ yields increased mutual information between network input and ground-truth label, formally:
\begin{equation}
\label{eq:prop3_0}
    I([t_{\mathrm{CLS}}^{\mathrm{Em}}+\delta,P_0\mathbf{T}^{\mathrm{Em}}_\mathrm{img}];Y)>I([t_{\mathrm{CLS}}^{\mathrm{Em}},P_0\mathbf{T}^{\mathrm{Em}}_\mathrm{img}];Y)
\end{equation}
\end{proposition}
Refer to the supplementary material for detailed proof.


\cref{prop:3} indicates that optimizing the augmentation vector $\delta$ on [CLS] effectively compensates for the mutual information loss introduced by token aggregation. 
Here we leverage [CLS] embedding optimization to augment its information. 
Specifically, for each image batch $\mathbf{x}$ during the TTA process, we directly update the augmentation vector $\delta$ using gradients derived from the optimization objective $\mathcal{L}(\theta;\mathbf{x})$:
\begin{equation}
    \widetilde{\delta}=\delta-\eta_\delta\nabla_\delta\mathcal{L}(\theta;\mathbf{x}),
\end{equation}
where $\eta_\delta$ denotes the learning rate for $\delta$, and $\nabla_\delta$ the gradient operation with respect to $\delta$.

\noindent\textbf{Shallow Layer [CLS] Bias Augmentation.}
Simply learning an augmentation vector $\delta$ for the [CLS] token has shown effectiveness in compensating for information loss (see \cref{sec:experiments} for details).
However, since the augmentation vector only operates on the embedding layer, it has limited flexibility to handle the information loss in deeper layers. 
As token aggregation progressively reduces the total number of tokens, information loss also accumulates gradually throughout the forward pass.
Consequently, it is necessary to develop a gradual information augmentation approach to compensate this incremental loss.

A straightforward approach is to learn an augmentation vector $\delta_l$ for the [CLS] token at every layer $l$, enabling each $\delta_l$ to compensate for the information loss at its corresponding layer.
However, this approach significantly increases the number of trainable parameters to $Ld$, where $L$ is the total number of layers in the ViT model. 
Prior TTA research \cite{cotta}
indicates that increasing trainable parameters complicates the optimization process and adds computational overhead.
In the ETTA scenario where computation budget is strictly limited, 
a large number of trainable parameters can cause suboptimal optimization and consequently degrade performance.

To achieve an optimal balance between efficiency and performance, we propose to augment [CLS] biases only in a selected subset $\mathbf{L}_{\mathrm{aug}}\subset\mathbf{L}$, with a predefined budget constraint $|\mathbf{L}_\mathrm{aug}|=L_{\mathrm{bgt}}$.
According to \cref{prop:3}, augmented biases compensate for the mutual information loss at their respective layers. 
Empirical evidence suggests that different ViT layers encode distinct features (e.g., color or texture) \cite{dorszewski2025colors}, leading to varied functional roles for the [CLS] token.
Ideally, the [CLS] features across \textit{all layers} should maintain high mutual information with the predictions, represented as maximizing $\sum _{l=1}^LI(t^l_{\mathrm{CLS}};Y)$.
Let $\Delta I_l$ denote the information compensation at layer $l$ resulting from [CLS] bias tuning at that layer.
As compensation introduced at layer $l$ propagates through subsequent $L-l-1$ layers, its total information compensation becomes $\Delta I_l(L-l)$. Hence, 
the optimization objective is to maximize the cumulative mutual information compensation from tuning across all selected layers
$\mathbf{L}_{\mathrm{aug}}$: 
\begin{equation}
    \max\ \sum_{l\in \mathbf{L}_{\mathrm{aug}}}\Delta I_l(L-l).
\end{equation}

In the ViT structure, the [CLS] tokens share the same hidden dimension $d$ across layers. 
In practice, a small augmentation bias consistently applied across layers yields approximately uniform compensation. We denote that uniform compensation value as $I_\mathrm{aug}$, and $\Delta I_l \simeq I_\mathrm{aug}$.
Thus, we formalize the selection of $\mathbf{L}_{\mathrm{aug}}$ as an optimization problem:
\begin{equation}
\label{eq:subset}
\begin{aligned}
\arg \max_{\mathbf{L}_{\mathrm{aug}}} &\sum_{l\in\mathbf{L}_{\mathrm{aug}}}I_{\mathrm{aug}}(L-l) \\
\mathrm{s.t.}&\;|\mathbf{L}_\mathrm{aug}|=L_{\mathrm{bgt}}
\end{aligned}
\end{equation}
The solution to \cref{eq:subset} is straightforwardly determined as:
\begin{equation}
\label{eq:subset_solution}
    \mathbf{L}_{\mathrm{aug}}=\{0,1,\dots,L_\mathrm{bgt}-1\},
\end{equation}
which indicates that augmenting [CLS] biases in shallower layers provides the maximal total mutual information compensation. 
Adopting this solution, we specifically update the augmentation vector $\delta_l$ for each $l\in \mathbf{L}_{\mathrm{aug}}$ using the gradient derived from the optimization objective $\mathcal{L}(\theta;\mathbf{x})$:
\begin{equation}
    \widetilde{\delta_l}=\delta_l-\eta_{\delta_l}\nabla_{\delta_l}\mathcal{L}(\theta;\mathbf{x}).
\end{equation}
where $\eta_{\delta_l}$ is the learning rate applied across selected layers.

\noindent\textbf{Overall TTA Framework.}
The proposed information augmentation strategies are implemented with the TTA framework, constitute our ETTA method, NAVIA.
For the overall learning objective, we follow recent established approaches \cite{FOA,MGTTA,PromptAlign}, empolying entropy minimization and feature statistics discrepancy loss. 
Specifically, to calculate the above discrepancy loss, following \cite{FOA,PromptAlign}, we first sample a small set (e.g., 64) of unlabeled images from the source domain to compute the mean feature $\{\mu_l^S\}_{l=1}^L$ and mean variance $\{\sigma_l^S\}_{l=1}^L$ of each layer $l$. During TTA, we calculate the corresponding mean $\{\mu_l(\mathbf{T}_l)\}_{l=1}^L$ and variance $\{\sigma_l(\mathbf{T}_l)\}_{l=1}^L$ of target domain samples and align these with the source statistics. Formally, the loss is defined as:
\begin{flalign}
\label{eq:tta_loss}
     &\mathcal{L}_{TTA} =  \sum_{c \in \mathcal{C}} -\hat{y}_c \log \hat{y}_c + \nonumber \\ 
     &\lambda \sum_{l=1}^{L} (\parallel \mu_l(\mathbf{T}_l)-\mu_l^S \parallel^2 + \parallel \sigma_l(\mathbf{T}_l) - \sigma_l^S \parallel^2),
\end{flalign}
where $\mathcal{C}$ denotes the set of categories, $\hat{y}_c$ is the probability prediction for category $c$, and $\lambda$ is a trade-off parameter.
During the TTA process, each incoming test batch is processed by a single forward pass before performing back-propagation. 
Importantly, the predictions for each batch are obtained \textit{before} back-propagation to ensure computational efficiency, with only one forward pass per batch during TTA.
For ETTA, NAVIA applies ToMe \cite{ToMe} during the forward pass, and leverages information augmentation of the [CLS] embedding and shallow-layer [CLS] biases during back-propagation. Following previous TTA methods, NAVIA also adopts standard norm-tuning for basic distribution alignment. Experiment results demonstrate that NAVIA enhances both model adaptation efficiency and accuracy.
\section{Experiments}
\label{sec:experiments}

\begin{table*}[t!]
  \centering
  \resizebox{\textwidth}{!}{\begin{tabular}{l | c c c | c c c c | c c c c | c c c c | c | c}
    \toprule
    & \multicolumn{3}{c}{Noise} & \multicolumn{4}{c}{Blur} & \multicolumn{4}{c}{Weather} & \multicolumn{4}{c}{Digital} & & \\
    \textbf{Method} & Gauss. & Shot & Imp. & Def. & Glass
        & Mot. & Zoom & Snow & Frost & Fog
        & Brit. & Cont. & Elas. & Pix. & JPEG & Avg. & GFLOPs \\
    \midrule
    NoAdapt & 56.8 & 56.8 & 57.5 & 46.9 & 35.6 & 53.1 & 44.8 & 62.2 & 62.5 & 65.7 & 77.7 & 32.6 & 46.0 & 67.0 & 67.6 & 55.5 & 15.71 \\
    \midrule
    Tent & 60.3 & 61.6 & 61.8 & 59.2 & 56.5 & 63.5 & 59.2 & 54.3 & 64.5 & 2.3 & 79.1 & 67.4 & 61.5 & 72.5 & 70.6 & 59.6 & 15.71 \\
    LAME & 56.5 & 56.5 & 57.2 & 46.4 & 34.7 & 52.7 & 44.2 & 58.4 & 61.5 & 63.1 & 77.4 & 24.7 & 44.6 & 66.6 & 67.2 & 54.1 & 15.71\\
    CoTTA & 63.6 & 63.8 & 64.1 & 55.5 & 51.1 & 63.6 & 55.5 & 70.0 & 69.4 & 71.5 & 78.5 & 9.7 & 64.5 & 73.4 & 71.2 & 61.7 & 31.42 \\
    SAR & 59.2 & 60.5 & 60.7 & 57.5 & 55.6 & 61.8 & 57.6 & 65.9 & 63.5 & 69.1 & 78.7 & 45.7 & 62.4 & 71.9 & 70.3 & 62.7 & 31.42 \\
    EATA & 61.2 & 62.3 & 62.7 & 60.0 & 59.2 & 64.7 & 61.7 & 69.0 & 66.6 & 71.8 & 79.7 & 66.8 & 65.0 & 74.2 & 72.3 & 66.5 & 15.71 \\
    DeYO & 62.4 & 64.0 & 63.9 & 61.0 & 60.7 & 66.4 & 62.9 & 70.9 & 69.6 & 73.7 & 80.5 & 67.2 & 69.9 & 75.7 & 73.7 & 68.2 & 31.42 \\
    FOA & 61.5 & 63.2 & 63.3 & 59.3 & 56.7 & 61.4 & 57.7 & 69.4 & 69.6 & 73.4 & 81.1 & 67.7 & 62.7 & 73.9 & 73.0 & 66.3 & 446.49 \\
    \midrule
    ToMe$_{4}$ & 63.3 & 65.3 & 65.1 & 62.3 & 63.1 & 68.0 & 67.0 & 73.2 & 70.8 & 75.6 & 81.3 & 68.4 & 73.0 & 77.0 & 74.9 & 69.9 & 14.29 \\
    ToMe$_{8}$ & 62.0 & 63.5 & 63.4 & 61.0 & 62.2 & 67.1 & 66.0 & 72.5 & 69.8 & 74.8 & 81.0 & 67.0 & 72.0 & 76.3 & 74.3 & 68.9 & \textbf{12.22} \\
    EViT$_{4}$ & 62.7 & 64.4 & 64.1 & 61.8 & 62.5 & 67.4 & 65.6 & 72.2 & 70.4 & 73.8 & 80.9 & 68.5 & 71.8 & 76.2 & 73.6 & 69.1 & 14.80 \\
    EViT$_{8}$ & 59.9 & 61.8 & 61.6 & 60.2 & 60.8 & 65.6 & 63.8 & 70.4 & 68.2 & 72.3 & 79.8 & 66.2 & 70.5 & 74.6 & 71.6 & 67.2 & 12.73 \\
    Tofu$_{4}$ & 63.5 & 64.9 & 65.1 & 61.9 & 62.7 & 67.7 & 66.6 & 73.0 & 70.6 & 75.7 & 81.1 & 68.6 & 72.6 & 76.8 & 74.6 & 69.7 & 14.29 \\
    Tofu$_{8}$ & 62.2 & 63.9 & 63.7 & 60.9 & 61.8 & 66.8 & 65.4 & 72.1 & 69.6 & 74.4 & 80.7 & 67.3 & 71.8 & 76.2 & 74.0 & 68.7 & \textbf{12.22} \\
    \midrule
    NAVIA$_{4}$ & \textbf{64.2} & \textbf{65.9} & \textbf{65.7} & \textbf{63.5} & \textbf{64.4} & \textbf{69.0} & \textbf{67.4} & \textbf{74.0} & \textbf{72.0} & \textbf{76.9} & \textbf{81.4} & \textbf{69.8} & \textbf{73.8} & \textbf{77.4} & \textbf{75.5} & \textbf{70.7} & 14.29 \\
    NAVIA$_{8}$ & 62.7 & 64.4 & 64.2 & 62.1 & 63.5 & 67.9 & 66.5 & 73.4 & 71.1 & 76.4 & 81.0 & 68.8 & 73.1 & 76.7 & 74.9 & 69.8 & \textbf{12.22} \\
    \bottomrule
  \end{tabular}}
  \caption{Performance (\%) comparison with state-of-the-art methods on ImageNet-C (severity level 5). Subscripts for method names represent the number of tokens reduced in each layer through token aggregation. \textbf{Bold} represents the best result. }
  \label{tab:corruption_full}
\end{table*}

\begin{table*}[htbp]
\centering
\resizebox{\textwidth}{!}{\begin{tabular}{l|ccc|c|l|ccc|c|l|ccc|c}
\toprule
Method & R & V2 & S & Avg. & Method & R & V2 & S & Avg. & Method & R & V2 & S & Avg. \\
\midrule
NoAdapt & 59.5 & 75.4 & 44.9 & 59.9 & DeYO    & 66.1 & 74.8 & 52.2 & 64.4 & EViT$_8$ & 64.4 & 73.8 & 49.9 & 62.7 \\
Tent   & 63.9 & 75.4 & 49.9 & 63.0 & FOA     & 63.8 & 75.4 & 49.9 & 63.0 & Tofu$_4$ & 67.3 & 75.3 & 52.1 & 64.9 \\
CoTTA  & 63.5 & 75.4 & 50.0 & 62.9 & ToMe$_4$& 67.6 & 75.3 & 52.2 & 65.0 & Tofu$_8$ & 66.8 & 74.9 & 51.6 & 64.4 \\
SAR    & 63.3 & 75.1 & 48.7 & 62.4 & ToMe$_8$& 66.8 & 74.6 & 52.0 & 64.5 & NAVIA$_4$& \textbf{68.8} & \textbf{75.5} & \textbf{53.6} & \textbf{66.0} \\
EATA   & 63.3 & 0.6  & 50.9 & 38.3 & EViT$_4$& 66.0 & 74.8 & 51.3 & 64.0 & NAVIA$_8$& 68.4 & 75.2 & 53.4 & 65.6 \\
\bottomrule
\end{tabular}}
\caption{Performance (\%) comparison with state-of-the-art methods on ImageNet-R/V2/Sketch using ViT-Base. Subscripts for method names represent the number of tokens reduced in each layer through token aggregation. \textbf{Bold} represents the best result.}
\label{tab:rvs_full}
\end{table*}

\subsection{Experimental Setting}

\noindent\textbf{Datasets\&Models.}
Following \cite{FOA}, we evaluate our approach on four commonly used OOD benchmarks: ImageNet-C \cite{imagenetc}, ImageNet-R \cite{imagenetr}, ImageNet-V2 \cite{imagenetv2}, and ImageNet-S \cite{imagenets} with ViT-B/16. 

\noindent\textbf{Baselines.}
We compare NAVIA against: 
(1) \textit{Online TTA methods}, including Tent \cite{Tent}, LAME \cite{LAME}, CoTTA \cite{cotta}, SAR \cite{SAR}, FOA \cite{FOA} 
, EATA \cite{EATA}, and DeYO \cite{DeYO}, which follow an online batch-processing pipeline and predict on streamingly test batches.
We used their implementations from \cite{FOA};
(2) \textit{Token aggregation methods}, including ToMe \cite{ToMe}, EViT \cite{EViT}, and Tofu \cite{Tofu}, perform training-free token reduction within the ViTs, as defined in \cref{sec:prelim}.
For fair comparison under TTA, we implement these methods \textit{within our TTA framework} stated in \cref{sec:our_method}, with consistent token aggregation scheduling by reducing $r$ tokens in each layer. Experiments are conducted at two compression levels: moderate ($r=4$) and high ($r=8$).

\noindent\textbf{Implementation Details.}
Following \cite{FOA}, we set the learning rate to 5e-3, the batch size to 64, and the trade-off parameter $\lambda$ as 30.
We use SGD for both LayerNorm tuning and information augmentation. 
For token aggregation, we adopt ToMe \cite{ToMe} for NAVIA, given its superior baseline performance. 
All latency measurements are conducted using a GeForce RTX 3090 GPU. For more details, please refer to the supplementary material. 

\begin{table}[t]
\centering
\resizebox{\columnwidth}{!}{\begin{tabular}{c|cc|cccc|c}
\toprule
$r$ & [CLS]-E & [CLS]-S & C & R & V2 & Sketch & Avg. \\
\midrule
\multirow{4}{*}{4} & $\times$ & $\times$ & 69.9 & 67.6 & 75.3 & 52.2 & 66.3 \\
&$\surd$ & $\times$ & 70.3 & 68.2 & 75.2 & 52.6 & 66.6 \\
&$\times$ & $\surd$ & 70.4 & 68.0 & \textbf{75.5} & 53.4 & 66.8 \\
&$\surd$ & $\surd$ & \textbf{70.7} & \textbf{68.8} & \textbf{75.5} & \textbf{53.6} & \textbf{67.2} \\
\midrule
\multirow{4}{*}{8} &$\times$ & $\times$ & 68.9 & 66.8 & 74.6 & 52.0 & 65.6 \\
&$\surd$ & $\times$ & 69.3 & 67.5 & 75.0 & 52.9 & 66.2 \\
&$\times$ & $\surd$ & 69.5 & 67.3 & 74.7 & 53.0 & 66.1 \\
&$\surd$ & $\surd$ & \textbf{69.8} & \textbf{68.4} & \textbf{75.2} & \textbf{53.4} & \textbf{66.7} \\
\bottomrule
\end{tabular}}
\caption{Ablation study performance (\%) on four OOD benchmarks. [CLS]-E and [CLS]-S abbreviate [CLS] embedding augmentation and shallow layer [CLS] bias augmentation, respectively. \textbf{Bold} represents the best result. }
\label{tab:ablation}
\end{table}

\subsection{Main Results}

\noindent\textbf{Performance on ImageNet-C.} 
As shown in \cref{tab:corruption_full}, even integrated with token aggregation, 
NAVIA is the best performing TTA method at both compression rates.
NAVIA also outperforms all token aggregation methods when applied at the same compression rate. 
Notably, at the moderate compression rate, NAVIA achieves the best results for all corruptions. At the high compression rate, it performs comparably to the best moderate-compressed baseline.

\noindent\textbf{Performance on ImageNet-R/V2/S.}
In \cref{tab:rvs_full}, NAVIA consistently surpasses all TTA baselines across both compression rates. Even under the high compression rate, NAVIA surpasses all baselines, well demonstrating its superiority.

\noindent\textbf{Computational Efficiency.} 
As shown in \cref{fig:teaser}, 
NAVIA yields the best Pareto frountier over all methods, with the best accuracy among the most efficient methods.
Unlike other token aggregation techniques that suffer severe performance drop at higher compression rates, our method only experiences a minimal performance degradation (-0.9\%) when increasing the compression rate from $r=4$ (12.5\% sparsity) to $r=8$ (25\% sparsity).

Overall, these results validate that NAVIA achieves state-of-the-art performance in both accuracy and computational efficiency, successfully addressing the ETTA challenge.

\subsection{Experimental Analysis}

\noindent\textbf{Ablation Studies for Components.}
We do controlled experiments to analyze the contribution of each component in NAVIA: [CLS] embedding augmentation (``$[\mathrm{CLS}]$ Embed.'') and shallow layer [CLS] bias augmentation (``$[\mathrm{CLS}]$ Shallow''). 
As shown in \cref{tab:ablation}, both the two components independently improve the model’s adaptation performance. 
Notably, employing both strategies yields the best results across all experimental settings, underscoring their individual effectiveness and mutual complementarity.

\begin{table}[t]
\centering
\resizebox{\columnwidth}{!}{\begin{tabular}{c|c|cccc|c}
\toprule
Method & Subset & Noise & Blur & Wea. & Dig. & Avg. \\
\midrule
Shallow & \{0,1,2,3,4,5\} & \textbf{63.7} & \textbf{65.0} & \textbf{75.5} & \textbf{73.4} & \textbf{69.8} \\
Deep & \{6,7,8,9,10,11\} & 63.5 & 64.6 & 74.4 & 72.8 & 69.2 \\
Uniform & \{0,2,4,6,8,10\} & 63.4 & 64.5 & 74.6 & 72.8 & 69.2 \\
\bottomrule
\end{tabular}}
\caption{Performance (\%) analysis of augmenting [CLS] bias on different layers. Wea.\ and Dig.\ abbreviate Weather and Digital, respectively. \textbf{Bold} represents the best result. }
\label{tab:layer_method}
\end{table}

\begin{table}[t]
\centering
\setlength{\tabcolsep}{4.8pt}
\begin{tabular}{c|cccc|c}
\toprule
Method & Noise & Blur & Wea. & Dig. & Avg. \\
\midrule
Init. [CLS] Embed. & 60.2 & 60.8 & 72.0 & 69.3 & 65.9 \\
Add New Tokens & 63.6 & 64.7 & 74.7 & 73.0 & 69.4 \\
Tune Trained [CLS] & \textbf{63.7} & \textbf{65.0} & \textbf{75.5} & \textbf{73.4} & \textbf{69.8} \\
\bottomrule
\end{tabular}
\caption{Performance (\%) analysis of different token tuning choice approaches for [CLS] embedding augmentation. Wea.\ and Dig.\ abbreviate Weather and Digital, respectively. \textbf{Bold} represents the best result. }
\label{tab:token_method}
\end{table}

\begin{figure}[t]
    \centering
    \includegraphics[width=0.85\linewidth]{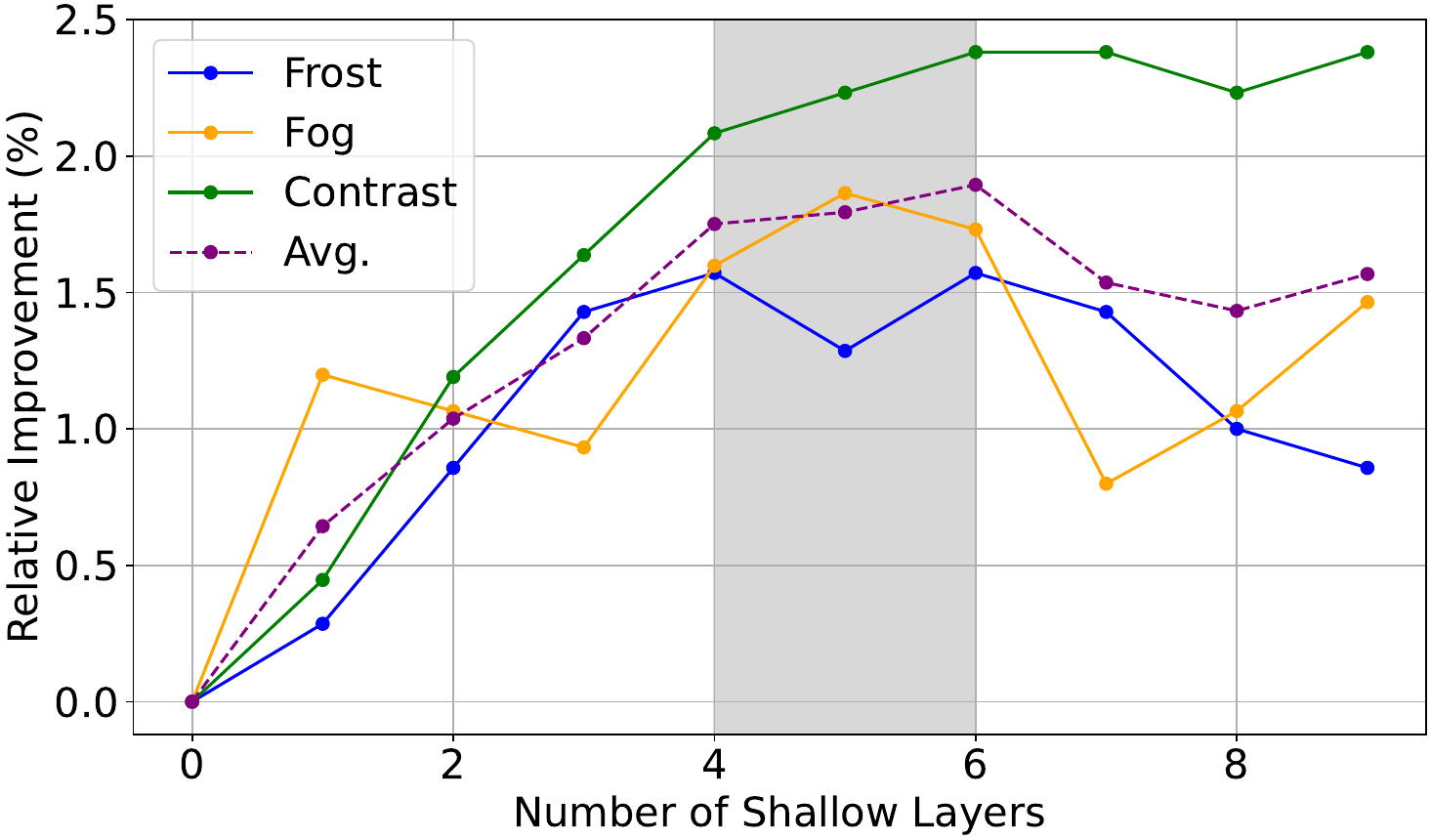}
    \caption{Relative acc. (\%) improvement results for different number of shallow layers with augmented [CLS] biases.}
    \label{fig:layer_choice}
\end{figure}

\noindent\textbf{Design Choices for NAVIA.}
We study the two design factors of [CLS] bias augmentation: 
the selection of layers for tuning and the choice for token augmentation.
(1) \textit{Layer Selection.} 
We investigate layer subsets where [CLS] biases are augmented.
First, as shown in \cref{tab:layer_method}, compared to other two selections (i.e., deep layers, and uniformly selected layers), augmenting [CLS] biases exclusively in shallow layers consistently yields the best results. 
Second, considering the number of shallow layers, as shown in \cref{fig:layer_choice}, tuning 4 to 6 shallow layers is optimal. 
Extending augmentation to more than 6 layers leads to performance saturation or slight degradation, likely due to suboptimal optimization. 
This can be attributed to the limited number of iterations available during the TTA process. Introducing an excessive number of tunable parameters hinders the convergence of all bias parameters. As a result, the optimization process becomes less effective, which ultimately impairs performance.
Consequently, we select the optimal number of layers from the set \{4,5,6\} based on minimizing the error computed using \cref{eq:tta_loss} on a small set of gathered target images. 
(2) \textit{Token Selection.}
We then compare our proposed pre-trained [CLS] tuning strategy against alternative token augmentation methods.
As shown in \cref{tab:token_method}, tuning newly added tokens leads to lower performance, likely due to the difficulty of tuning from scratch with limited TTA steps. 
Tuning re-initialized [CLS] notably decreases performance, highlighting the importance of preserving pre-trained embeddings that inherently encode class-relevant information.

\begin{figure}[!t]
    \centering
    \includegraphics[width=0.95\linewidth]{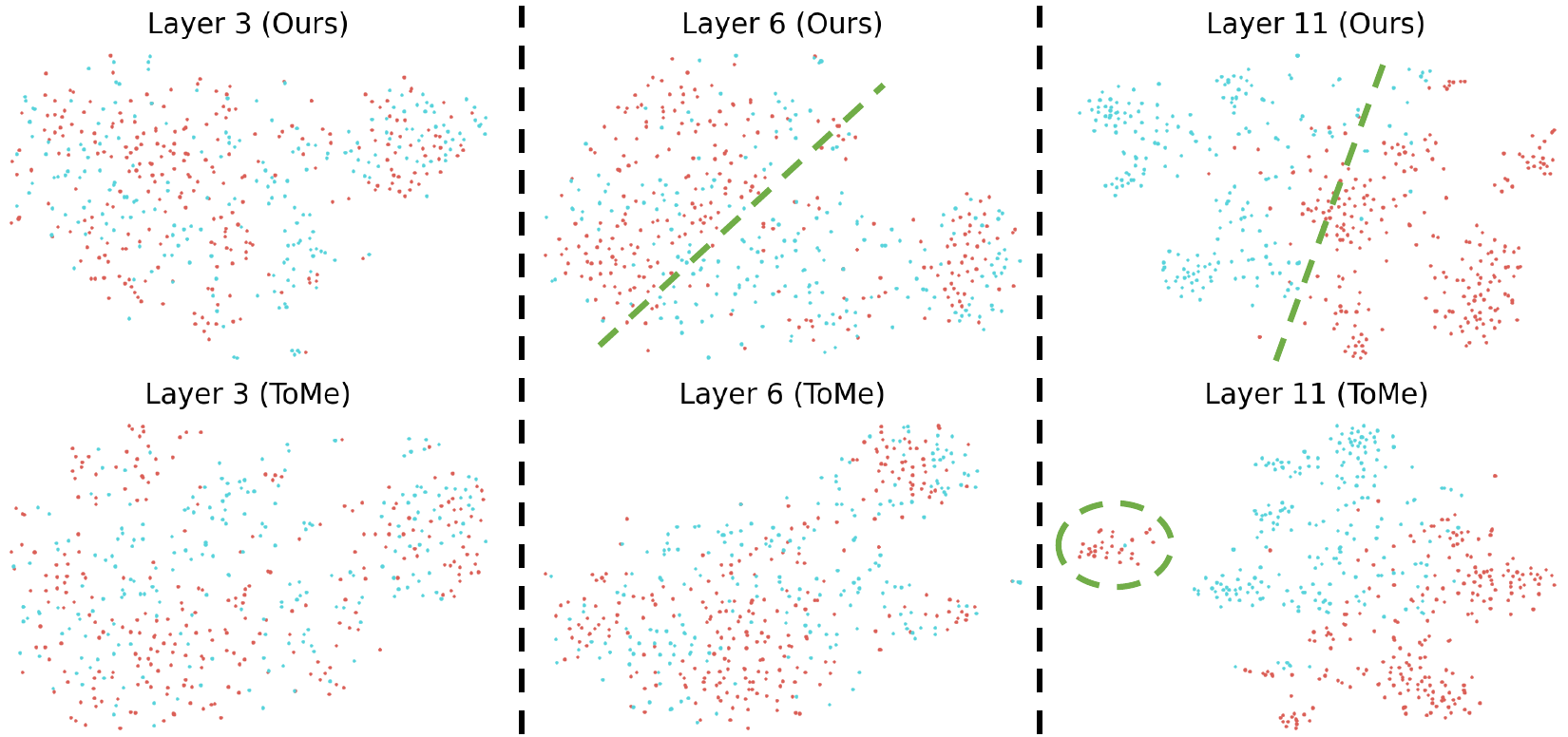}
    \caption{t-SNE analysis results. Colors represent different super-classes (aquatic animals in red, birds in blue).}
    \label{fig:tsne}
\end{figure}

\noindent\textbf{Visualizations.}
We conduct a qualitative analysis to illustrate how NAVIA affects feature representations.
Specifically, we select 8 representative classes from ImageNet-R, categorized into two ``super-classes'': aquatic animals (goldfish, great white shark, hammerhead, stingray) and birds (hen, ostrich, goldfinch, junco). We apply t-SNE \cite{tsne} to the [CLS] features extracted from layers 3, 6, and 11 of ViT.
As shown in \cref{fig:tsne}, 
NAVIA progressively separates features between the two super-classes from shallow to deep layers, while effectively preserving the intra-class distributions.
Conversely, with ToMe, the [CLS] features from different super-classes in layers 3 and 6 exhibit significant overlap, indicating limited discriminability.
Even in layer 11, ToMe seems to misclassify a class to the other super-classes. 
These results underscore that our proposed information augmentation method guides the ViT to better capture class-specific semantics under distribution shifts, resulting in improved adaptation performance.

\begin{figure}
    \centering
    \includegraphics[width=0.85\linewidth]{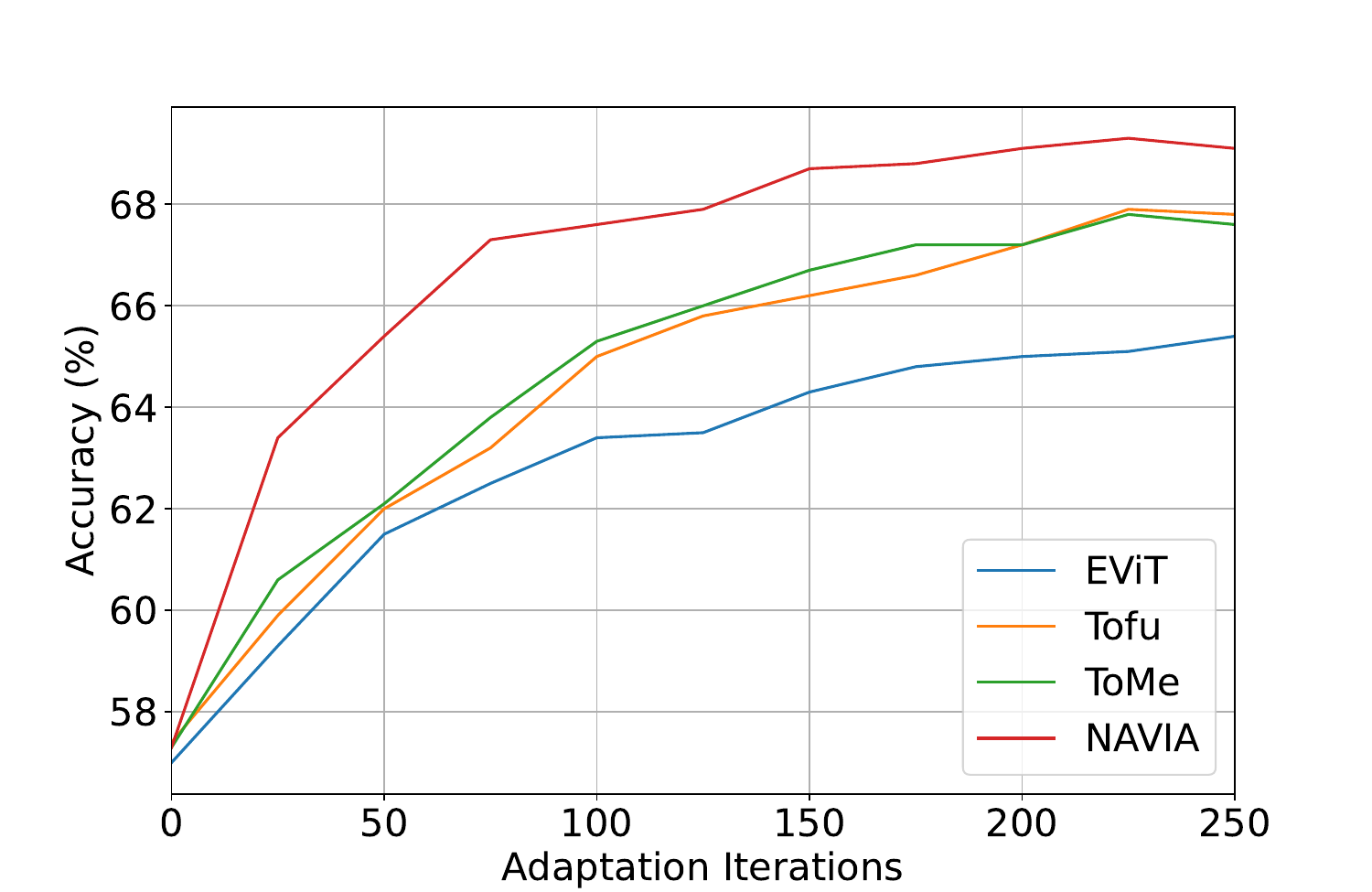}
    \caption{Convergence speed comparison on ImageNet-R.}
    \label{fig:convergence}
\end{figure}

\noindent\textbf{Convergence Speed Analysis.}
To analyze the convergence efficiency of NAVIA, we compare it against existing token aggregation baselines following the evaluation protocol established by \cite{MGTTA}. Specifically, we perform TTA optimization over a fixed number of iterations and subsequently evaluate the optimized model’s average accuracy on a set of 10,000 images reserved from the original ImageNet-R dataset. As illustrated in \cref{fig:convergence}, NAVIA consistently achieves the fastest convergence rate and highest final accuracy among all compared methods. Notably, NAVIA demonstrates a rapid accuracy improvement in the initial adaptation iterations, indicating that our proposed information augmentation strategies effectively mitigate the early-stage information loss incurred by token aggregation.

\begin{figure}
    \centering
    \includegraphics[width=1.0\linewidth]{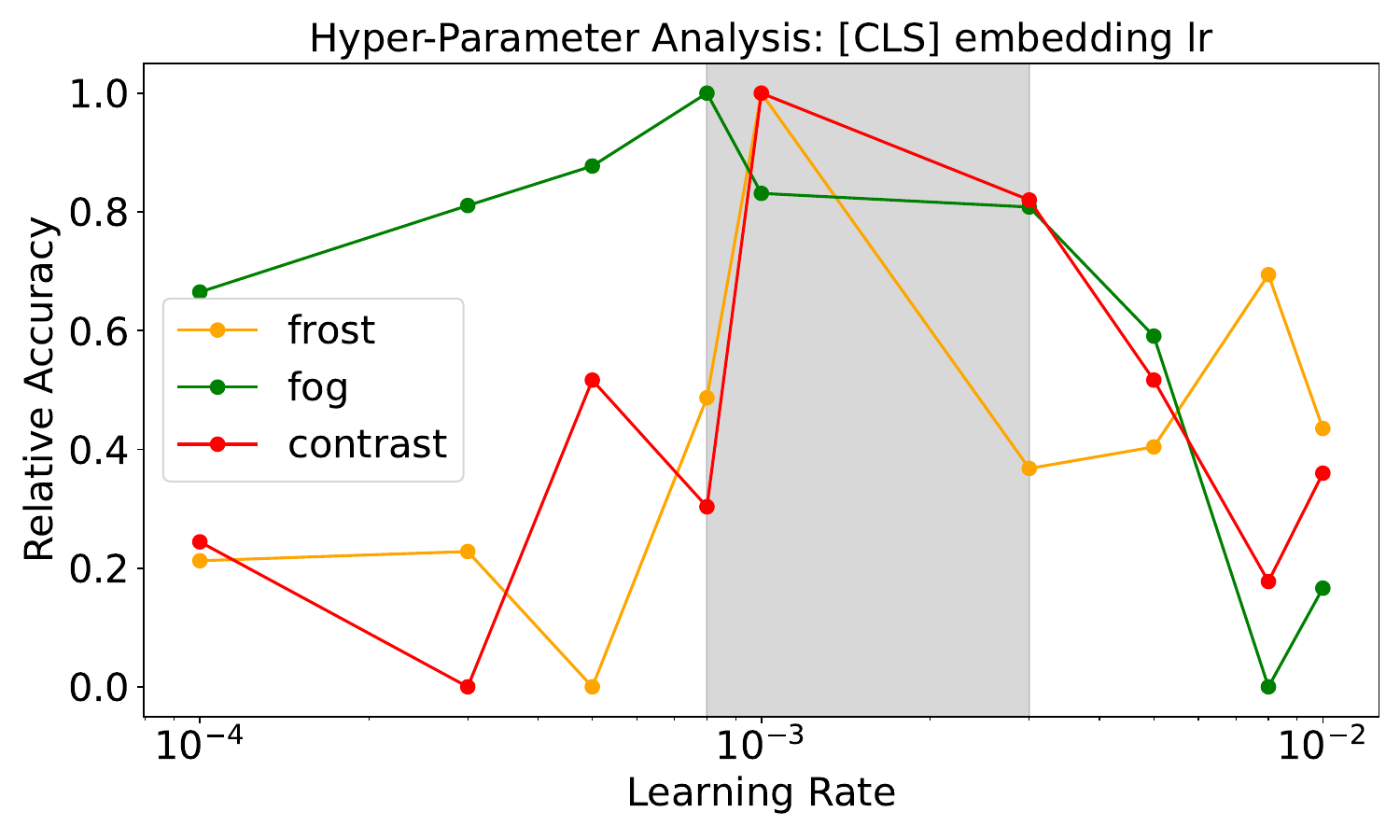}
    \caption{Hyper-parameter analysis of learning rates for the [CLS] embedding augmentation. Relative accuracy values are plotted for the frost, fog, and contrast corruption. The best-performing learning rates are highlighted in grey.}
    \label{fig:param_embed}
\end{figure}

\begin{figure}
    \centering
    \includegraphics[width=1.0\linewidth]{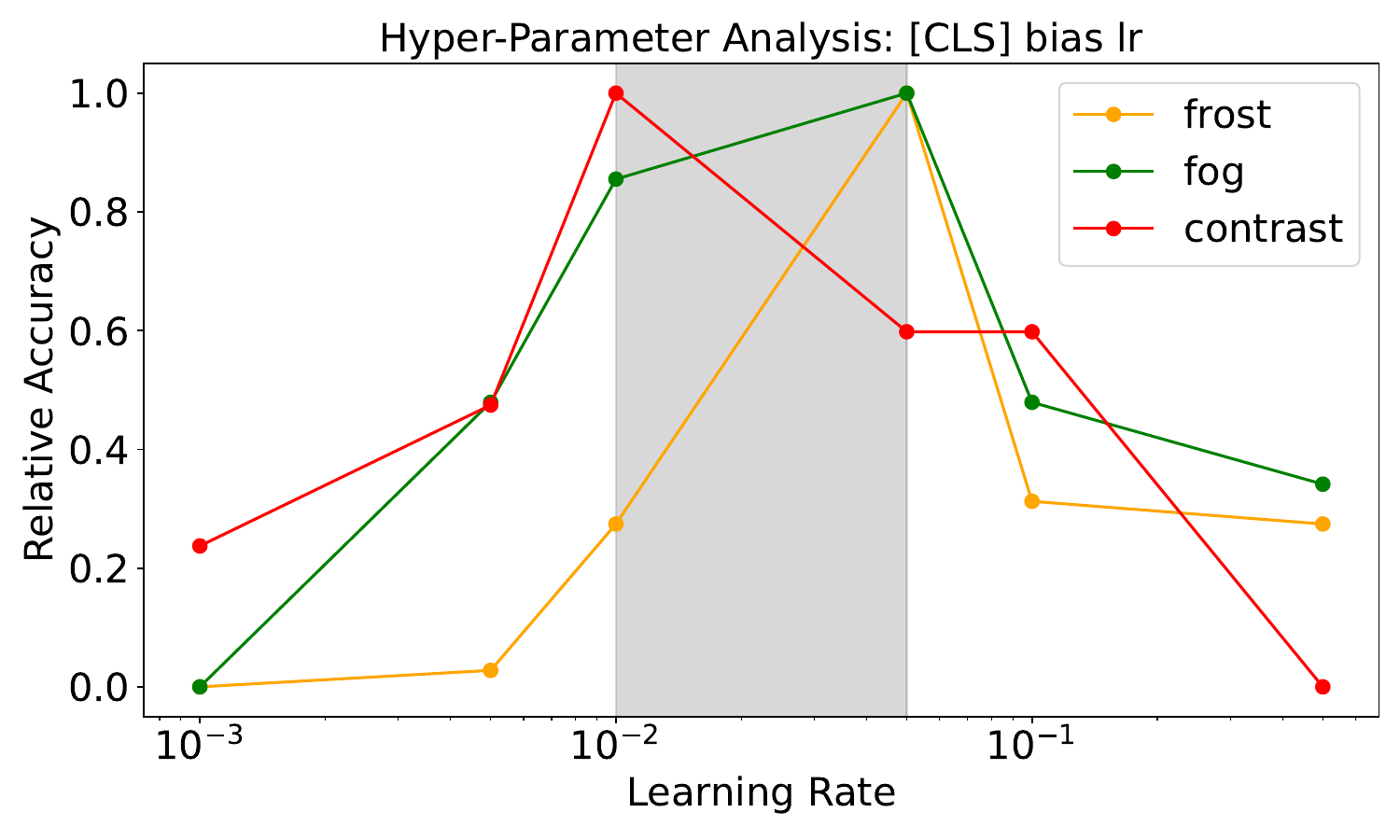}
    \caption{Hyper-parameter analysis of learning rates for the [CLS] bias augmentation. Relative accuracy values are plotted for the frost, fog, and contrast corruption. The best-performing learning rates are highlighted in grey.}
    \label{fig:param_bias}
\end{figure}

\noindent\textbf{Hyperparameter Sensitivity Analysis.}
We conduct sensitivity analysis for the learning rate of  the [CLS] embedding $\eta_\delta$ and the learning rate of the shallow layer [CLS] biases $\eta_{\delta_l}$. Specifically, we modify each hyperparameter with all other hyperparameter fixed, and report the change of OOD performance on different modified values for the current hyperparameter. 
As each dataset naturally exhibits different difficulties, to present the results clearer, we plot the ``relative accuracy'' to illustrate the trend of how accuracy changes. 
``Relative accuracy'' $\mathcal{A}_{\mathrm{rlt}}$ is defined as:
\begin{equation}
    \mathcal{A}_{\mathrm{rlt}} = \frac{\mathcal{A}_{\mathrm{act}}-\mathcal{A}_{\min}}{\mathcal{A}_{\max}-\mathcal{A}_{\min}},
\end{equation}
where $\mathcal{A}_{\min}$ and $\mathcal{A}_{\max}$ are the minimum and maximum accuracy on the dataset among the current hyperparameter span, and  $\mathcal{A}_{\mathrm{act}}$ represents the actual accuracy of current experiment.
As shown in \cref{fig:param_embed} and \cref{fig:param_bias}, the best OOD performance for each dataset mainly remains in a narrow range of learning rate settings. To ensure an easy and simple hyperparameter selection, we simply limit the learning rate range of all datasets as \{8e-4, 1e-3, 3e-3\} for $\eta_\delta$ and \{1e-2, 5e-2\} for $\eta_{\delta_l}$. Similar with choosing the number of shallow layers for bias augmentaion, for each dataset, we simply choose the learning rate that yields the lowest error (calculated by Eq. 7 in the main article) on a small set of gathered target images. 
\section{Conclusion}
\label{sec:conclusion}

In this work, we introduced a novel challenge termed Efficient Test-Time Adaptation (ETTA)  to address the trade-off between computational efficiency and predictive performance in existing TTA frameworks.
While token aggregation methods reduce computational overhead, our theoretical analysis reveals that they inherently induce information loss, which norm-tuning cannot compensate for, leading to significant performance degradation in TTA.
To address this, we proposed to Neutralize Token Aggregation via Information Augmentation (NAVIA), compensating for information loss by fine-tuning augmentation vectors on [CLS] token and adaptive biases in the shallow layers of ViTs. 
Extensive experiments validated that NAVIA achieves superior predictive accuracy while reducing computational costs, offering a practical solution for resource-constrained environments.

{
    \small
    \bibliographystyle{ieeenat_fullname}
    \bibliography{main}
}


\clearpage
\setcounter{section}{0}
\renewcommand{\thesection}{\Alph{section}} 

\section{More Implementation Details}

\subsection{Datasets}
To evaluate the effectiveness of our method under various types of distribution shifts, we adopt four widely-used out-of-distribution benchmarks: ImageNet-C \cite{imagenetc}, ImageNet-R \cite{imagenetr}, ImageNet-V2 \cite{imagenetv2}, and ImageNet-Sketch \cite{imagenets}. These datasets introduce diverse and challenging variations from the original ImageNet-1K distribution, providing a comprehensive testbed for assessing the robustness and adaptability of TTA methods.

\noindent\textbf{ImageNet-C} \cite{imagenetc} applies 15 types of generated corruptions—such as Gaussian noise, motion blur, fog, and JPEG compression—to the validation set of ImageNet-1K, each with five levels of severity. Following prior work, we use the highest severity level (level 5) to evaluate model performance under strong synthetic perturbations. This benchmark is designed to assess robustness to low-level signal distortions and is a standard benchmark for corruption robustness. For clarity, we adopt the standard abbreviations used in prior literature to denote each corruption type as presented in the main text table: Gaussian noise (Gauss.), shot noise (Shot), and impulse noise (Imp.) under the Noise category; defocus blur (Def.), glass blur (Glass), motion blur (Mot.), and zoom blur (Zoom) under Blur; snow (Snow), frost (Frost), fog (Fog), and brightness (Brit.) under Weather; and contrast (Cont.), elastic transformation (Elas.), pixelation (Pix.), and JPEG compression (JPEG) under Digital distortions.

\noindent\textbf{ImageNet-R(endition)} \cite{imagenetr} contains 30,000 images covering 200 ImageNet-1K classes, sourced from non-photographic domains such as art, cartoons, sketches, and 3D renderings. It introduces significant domain shift in visual style, testing the model’s ability to generalize to abstract representations and out-of-distribution textures and forms. The dataset emphasizes robustness to style variation beyond natural imagery.

\noindent\textbf{ImageNet-V2} \cite{imagenetv2} is a re-collection of ImageNet validation samples using a protocol intended to closely replicate the original data collection procedure, but with independently sourced images. Although label space and overall content are aligned with ImageNet-1K, ImageNet-V2 introduces subtle shifts in distribution that make it suitable for evaluating a model’s generalization capacity to near-domain but unseen samples.

\noindent\textbf{ImageNet-S(ketch)} \cite{imagenets} consists of black-and-white hand-drawn sketches aligned with 1,000 ImageNet-1K classes. The images are highly abstract and lack texture, color, and background cues, placing strong emphasis on shape-based recognition. This benchmark is particularly valuable for assessing a model's reliance on texture versus structure and its ability to adapt to minimal visual cues.

\section{Detailed Proof}

We leverage the Data-Processing Inequality (DPI) \cite{beaudry2012intuitive} for the theoretical analysis:
\begin{lemma}
    (DPI) Let three random variables $X$, $Y$, and $Z$ form a Markov chain $X\rightarrow Y\rightarrow Z$, no processing of $Y\rightarrow Z$ can increase the information that $Y$ contains about $X$: $I(X;Y)\ge I(X;Z)$.
\end{lemma}
In the theoretical analysis, we can treat an image $Y$ being dependent on its ground-truth label $X$, so any processing of the image by the ViT (i.e., $f(Y)=Z$) does not increase the information that the image contains about the ground-truth label.

\subsection{Proof for Prop. 3.2}

\begin{proof}

First, according to DPI, for any aggregation matrix $P_l$ incurred by the aggregation metric $g(\cdot)$, we have $I(P_l\mathbf{T}^{\mathrm{In}}_l;Y)\le I(\mathbf{T}^{\mathrm{In}}_l;Y)$. 
For any $P_l\in \mathbb{R}^{N_{l+1}\times N_l}$, where $N_{l+1}< N_l$, the matrix $P_l\mathbf{T}^{\mathrm{In}}_l$ has a smaller rank than $\mathbf{T}^{\mathrm{In}}_l$, resulting in the strict inequality $I(P_l\mathbf{T}^{\mathrm{In}}_l;Y)< I(\mathbf{T}^{\mathrm{In}}_l;Y)$. Since $t^l_{\mathrm{CLS}}$ is a weighted sum of image tokens, we can express $I(P_l\mathbf{T}^{\mathrm{In}}_l;Y)=I([t^l_{\mathrm{CLS}},P_l\mathbf{T}^{\mathrm{In}}_l];Y)\ge I(t^l_{\mathrm{CLS}};Y)$. 
Thus, after token aggregation, the [CLS] token in layer $l$ represents strictly less information than the image tokens:
\begin{equation}
    I(t^l_{\mathrm{CLS}};Y)<I(\mathbf{T}^{\mathrm{In}}_l;Y).
\end{equation}
According to DPI, tokens in subsequent layers of ViT do not contain more information than those in prior layers. Applying that through the forwarding process ($f_{\theta;g;\mathbf{x}}(\cdot)$) leads to that the [CLS] token \textit{output} represents strictly less information than the image token \textit{input}:
\begin{equation}
\label{eq:prop1_1}
    I(f_{\theta;g;\mathbf{x}}(t^{\mathrm{Em}}_{\mathrm{CLS}});Y)< I(\mathbf{T}^{\mathrm{Em}}_{\mathrm{Img}};Y).
\end{equation}
Considering that the ViT model has been pre-trained on a large image corpus \cite{vit} and validated on various downstream tasks \cite{Consolidator,PYRA}, 
we can reasonably assume that when forwarded by the original ViT (i.e., $f_{\theta;I;\mathbf{x}}(\cdot)$), the [CLS] token could still effectively represent the input image.
This leads to the following relationship:
\begin{equation}
\label{eq:prop1_2}
    I(f_{\theta;I;\mathbf{x}}(t^{\mathrm{Em}}_{\mathrm{CLS}});Y)=I(\mathbf{T}^{\mathrm{Em}}_{\mathrm{Img}};Y)
\end{equation}
Combining \cref{eq:prop1_1} and \cref{eq:prop1_2} yields the conclusion.
\end{proof}

\subsection{Proof for Prop. 3.3}

\begin{proof}
According to DPI, for any layer $l$, we have: 
\begin{equation}
\begin{aligned}
    I(f_{\theta;g;\mathbf{x}}(t^{\mathrm{Em}}_{\mathrm{CLS}});Y)&\le I(\mathrm{LN}_\pi([t^l_{\mathrm{CLS}},P_l\mathbf{T}^{\mathrm{In}}_l]);Y) \\
    &\le I([t^l_{\mathrm{CLS}},P_l\mathbf{T}^{\mathrm{In}}_l];Y)\\ &<I([t^l_{\mathrm{CLS}},\mathbf{T}^{\mathrm{In}}_l];Y)
\end{aligned}
\end{equation}
Here, the first and the second inequalities hold due to the DPI, as any processing operations inside the ViT cannot increase the information that the token sequence contains about the ground-truth. 
The strict inequality is guaranteed by Prop. 3.2, which demonstrates that applying token aggregation strictly decreases mutual information.
Therefore, for any possible affine parameters $\pi$ in the LayerNorm, the mutual information $I(\mathrm{LN}_\pi([t^l_{\mathrm{CLS}},P_l\mathbf{T}^{\mathrm{In}}_l]);Y)$ is strictly upper-bounded by $I([t^l_{\mathrm{CLS}},P_l\mathbf{T}^{\mathrm{In}}_l];Y)$, which itself is strictly smaller than $I([t^l_{\mathrm{CLS}},\mathbf{T}^{\mathrm{In}}_l];Y)$.
Consequently, tuning LayerNorm parameters cannot recover the mutual information loss induced by the token aggregation matrix $P_l$.
\end{proof}

\subsection{Proof for Prop. 3.4}

\begin{proof}
Consider the original input embedding consisting of the [CLS] embedding $t_{\mathrm{CLS}}^{\mathrm{Em}}$ and the image token embeddings $\mathbf{T}^{\mathrm{Em}}_\mathrm{img}=[t^{\mathrm{Em}}_1,t^{\mathrm{Em}}_2,\cdots,t^{\mathrm{Em}}_N]$. According to Prop. 3.2, at the first layer (indexed by $0$) of the ViT model, we have:
\begin{equation}
\label{eq:prop3_1}
I([t_{\mathrm{CLS}}^{\mathrm{Em}},\mathbf{T}^{\mathrm{Em}}_\mathrm{img}];Y)>I([t_{\mathrm{CLS}}^{\mathrm{Em}},P_0\mathbf{T}^{\mathrm{Em}}_\mathrm{img}];Y)
\end{equation}
In Prop. 3.3, we have shown that mutual information under LayerNorm optimization is upper-bounded by the right side of \cref{eq:prop3_1}, implying it cannot recover the information lost due to the token aggregation represented by $P_0$.

Now, define a learned augmentation bias vector $\delta$ on the [CLS] token embedding, modifying the [CLS] token input as $u=t^{\mathrm{Em}}_{\mathrm{CLS}}+\delta$. 
Then, the mutual information between the augmented [CLS] token input $u$ and the ground-truth label $Y$ can be expressed in terms of entropy as follows: 
\begin{equation}
\begin{aligned}
    I(u;Y)&=H(Y)-H(Y|u) \\
    &= H(Y)-H(Y;u)-H(u).\\
\end{aligned}
\end{equation}
Since $H(u)$ is an expectation over all possible $u$'s, it is a constant.
Thus, maximizing $I(u;Y)$ with respect to $\delta$ simplifies to minimizing the joint entropy $H(Y;u)$: 
\begin{equation}
\label{eq:prop3_2}
    \arg \max_\delta I(u;Y)=\arg \min_\delta H(Y;u).
\end{equation}
For the ViT model parameterized by $\theta$, used for image recognition, we explicitly model the posterior distribution as $q_\theta(Y|u)$. The joint distribution can then be decomposed into:
\begin{equation}
    p_\theta(Y;u)=p(u)q_\theta(Y|u).
\end{equation}
Consequently, the joint entropy can be rewritten as:
\begin{equation}
\begin{aligned}
H(Y;u)&=-\iint p_\theta(Y|u)\log(p_\theta(Y|u))\,dY\,du \\
&=-\iint p(u)q_\theta(Y|u) \log(p(u)q_\theta(Y|u))\,dy\,du \\
&=-\int p(u)\log(p(u))\left[\sum_Yq_\theta(Y|u)\right]\,du\\&\quad\quad-\mathbb{E}_{u\sim p(u)}\left[\sum_Yq_\theta(Y|u)\log(q_\theta(Y|u))\right] \\
&=H(u)+\mathbb{E}_{u\sim p(u)}\bigl[H(q_\theta(Y|u))\bigl].
\end{aligned}
\end{equation}
As established above, $H(u)$ is a constant. Hence, minimizing $H(Y;u)$ is equivalent to minimizing the conditional entropy of predictions given the input:
\begin{equation}
\label{eq:prop3_3}
    \arg \min_\delta H(Y;u)=\arg \min_\delta \mathbb{E}_{u\sim p(u)}\bigl[H(q_\theta(Y|u))\bigl].
\end{equation}
Combining \cref{eq:prop3_2} and \cref{eq:prop3_3}, we observe that entropy minimization on the augmentation bias vector $\delta$ effectively maximizes mutual information at the input level. Therefore, introducing and optimizing $\delta$ via entropy minimization yields mutual information greater than the initial condition where $\delta = 0$. This formally validates the inequality presented in Eq. 1 in the main article.
\end{proof}

\end{document}